\documentclass[journal]{IEEEtran}
\ifCLASSINFOpdf
\else
\fi

\usepackage{cite}
\usepackage{multirow}
\usepackage{amsmath}  
\usepackage{float}
\usepackage{booktabs} 

\usepackage{graphicx}
\usepackage[colorlinks,linkcolor=black]{hyperref}

\usepackage{bm}
\usepackage{lineno}
\usepackage{url}
\usepackage{amsmath, amsfonts}
\usepackage{tikz}

\usepackage{dsfont}
\usepackage{verbatim}
\usetikzlibrary{positioning}
\usepackage{calc}

\begin{document}
\flushbottom
%
\title{Graph-Stega: Semantic Controllable Steganographic Text Generation Guided by Knowledge Graph}
%
%
%

\author{Zhongliang~Yang,
        Baitao~Gong,
        Yamin~Li,
        Jinshuai~Yang,
        Zhiwen~Hu,
        Yongfeng~Huang,
\thanks{Z. Yang, J. Yang and Y. Huang are with the Department of Electronic Engineering, 
Tsinghua University, Beijing, 100084, China. E-mail: yfhuang@tsinghua.edu.cn}}

\maketitle

\begin{abstract}

Most of the existing text generative steganographic methods are based on coding the conditional probability distribution of each word during the generation process, and then selecting specific words according to the secret information, so as to achieve information hiding. Such methods have their limitations which may bring potential security risks. Firstly, with the increase of embedding rate, these models will choose words with lower conditional probability, which will reduce the quality of the generated steganographic texts; secondly, they can not control the semantic expression of the final generated steganographic text. This paper proposes a new text generative steganography method which is quietly different from the existing models. We use a Knowledge Graph (KG) to guide the generation of steganographic sentences. On the one hand, we hide the secret information by coding the path in the knowledge graph, but not the conditional probability of each generated word; on the other hand, we can control the semantic expression of the generated steganographic text to a certain extent. The experimental results show that the proposed model can guarantee both the quality of the generated text and its semantic expression, which is a supplement and improvement to the current text generation steganography.

\end{abstract}


%
\IEEEpeerreviewmaketitle

\section{Introduction}

Steganography is a research field with a long history of development. It mainly studies how to embed important information into common information carriers, hide the existence of it and thus to protect its security. Steganography has a very wide range of applications. In addition to providing military intelligence support, it can be used to protect the privacy of users in the communication process in daily life \cite{das2019enabling}, as well as copyright protection for digital media \cite{Seife917,Voss826,zhao1996watermarking}. Due to its extensive use, steganography has attracted wide attention of researchers in recent years.

The task of information hiding can be modeled as a non-cooperative game under the ``Prisoners' Problem" \cite{Simmons1984The,moulin2003information}, which can be briefly described as follows. Alice and Bob need to pass some secret information. However, all of their communication content must be reviewed by Eve. Once Eve detects that the transmitted information carrier contains secret information, the entire communication task fails. Basically, any information carrier that is not compressed to the Shannon limit has a certain degree of information redundancy, so they all can be used to embed additional information inside, such as digital medias like image\cite{fridrich2009steganography}, audio\cite{huang2011steganography}, text\cite{yang2018rnn} and video\cite{shanableh2012data}. Among them, text might have a higher information coding degree, which makes texts more difficult to embed additional information inside. However, text is one of the most commonly used communication carriers in people's daily life\cite{michel2011quantitative}. Therefore, studying how to effectively embed hidden information in texts has attracted a lot of researchers' interest\cite{Wayner1992Mimic,chapman1997hiding,moraldo2014approach,dai2010text,Luo2016Text,yang2018rnn,yang2018rits,fang2017generating,dai2019towards,ziegler2019neural}. Some early text steganography techniques were mainly implemented by making minor modifications to the text, such as adjusting word spacing\cite{Chotikakamthorn1998Electronic}, or by synonymizing specific words and phrases\cite{Xiang2014Linguistic}. These technologies are good attempts, but usually they can only have a very low embedding rate, so it is difficult for them to be practical\cite{yang2018rnn}. Therefore, later researchers began to try to use the text automatic generation model to achieve information embedding\cite{Luo2016Text,yang2018rnn,yang2018rits,fang2017generating,dai2019towards,ziegler2019neural}. 

The challenges for carrier generation steganography are obvious. Since the carrier will not be given in advance, the first challenge is: How to generate a semantically complete and natural enough steganographic sample? In order to solve this challenge, previous researchers have done a lot of works \cite{Luo2016Text,yang2018rnn,yang2018rits,fang2017generating,dai2019towards,ziegler2019neural}. From the first generation of steganography based on specific syntax structure \cite{chapman1997hiding}, to the later generation of steganography based on Markov model \cite{moraldo2014approach,shniperov2016text,yang2018automatically}, and the recent emergence of a method of steganography based on neural network\cite{yang2018rnn,yang2018rits,fang2017generating,dai2019towards,ziegler2019neural}, researchers have gradually been able to generate high-quality steganographic sentences. The current solution is mainly based on the following technical framework: Alice first uses a well-designed model (such as recurrent neural network\cite{yang2018rnn}) to learn a statistical language model of a large number of natural texts. Then Alice tries to encoded the conditional probability distribution of each word in the generation precess (Huffman coding \cite{yang2018rnn} or arithmetic coding \cite{ziegler2019neural}), so as to embed specific information into the generated natural sentences. This kind of methods seem to be able to solve the first challenge mentioned above, that is, generating natural enough steganographic texts. However, we found that there are still some defects, which may bring potential security risks.

Firstly, this kind of technical framework embeds information by encoding the conditional probability space of each word in the generated text. Therefore, with the increase of embedding rate, the model will be more and more likely to choose words with lower conditional probability, thus generating low-quality or even grammatical error steganographic sentences. Can we break this steganographic framework and select words with high conditional probability at any embedding rate, so as to ensure that the quality of the generated steganography text is always high enough? Secondly, the previous text generation steganographic methods usually can't control the semantic expression of the generated steganographic sentences. This may bring huge security risks. Z. Yang \emph{et al.} \cite{yang2019behavioral} have recently proposed a new security framework for covert communication systems. In this new framework, they find that if Alice does not control the semantics of the generated carrier and continues to generate a steganographic carrier with random semantics, even if the quality of each sentence is good enough, it may still bring other security risks of Alice and Bob's covert communication.

Trying to solve the two challenges of current text steganographic methods are the motivation of this paper. We abandon the current text steganography framework of ``language model + conditional probability coding" and try to use Knowledge Graph (KG) to guide the generation of steganographic sentences. On the one hand, we hide the secret information by coding the path in the knowledge graph, but not the conditional probability of each generated word, which is quitely different from previous text generation steganographic methods. On the other hand, we can use the knowledge Graph to guide the semantic expression of the generated steganographic text, so as to realize the generation of semantic controllable steganographic text to a certain extent. The scheme we proposed in this paper is probably not the most ideal solution to the above two challenges, but we hope to guide the follow-up researchers to create more diverse text generative steganography schemes, and to achieve semantic controllable text generative steganography as much as possible.

\section{Related Works}

Linguistic steganography based on automatic sentence generation technology has attracted a lot of researchers' attention for a long time due to its wide application prospects. In the early age, researchers could only generate sentences without semantic information and grammatical rules \cite{Wayner1992Mimic}. Later, some researchers tried to introduce syntactic rules to constrain the generated texts \cite{chapman1997hiding}, but the steganographic sentences generated by these methods were simple and could be easily recognized. After that, researchers tried to combine some natural language processing technologies to generate steganographic sentences \cite{moraldo2014approach,dai2010text,Luo2016Text,shniperov2016text,yang2018automatically}.

At present, most of the steganographic text automatic generation models are under the following framework: using a well-designed model to learn the statistical language model from a large number of normal sentences, and then implementing secret information hiding by encoding the conditional probability distribution of each word in the text generation process\cite{fang2017generating,yang2018automatically,yang2018rnn,yang2018rits,ziegler2019neural,dai2019towards}. In this framework, the early works mainly use Markov model to approximate the language model and calculate the conditional probability distribution of each word \cite{shniperov2016text,yang2018automatically}. However, due to the limitations of Markov model itself \cite{yang2018rnn}, the quality of the text generated by Markov model is still not good enough, which makes it easy to be recognized. In recent years, with the development of natural language processing technology, more and more steganographic text generation models based on neural network models have emerged\cite{fang2017generating,yang2018rnn,dai2019towards,ziegler2019neural,yang2018rits}. T. Fang \emph{et al.}\cite{fang2017generating} first divide the dictionary and fixedly encode each word, and then use the recurent neural network (RNN) to learn the statistical language model of natural text. Finally, in the automatic text generation process, different words are selected as output at each step according to the information that need to be hidden. Z. Yang \emph{et al.} \cite{yang2018rnn} also use a RNN to learn the statistical language model of a large number of normal sentences. Then they use a full binary tree (FLC) and a Huffman tree (VLC) to encode the conditional probability distribution of each word, and output corresponding words according to the information needs to be hidden, so as to realize the embedding of hidden information in the sentence generation precess. After that, Dai \emph{et al.} \cite{dai2019towards} and Ziegler \emph{et al.} \cite{ziegler2019neural} further improve the statistical language model and the coding method of conditional probability distribution, which can further optimize the conditional probability distribution of each word in the generated steganographic sentences.

These neural network based text automatic generation methods can better fit the statistical language model of normal sentences than Markov models, so that the quality of generated steganographic texts have been significantly improved \cite{fang2017generating,yang2018rnn,yang2018rits,ziegler2019neural,dai2019towards}. However, it still seems not enough. Z. Yang \emph{et al.}\cite{yang2019behavioral} recently proved through experiments that, due to these existing models cannot control the semantics of the generated steganographic sentences, even if the quality of the generated text is good enough, it will still bring potential security risks. Exploring the technology of automatic generation of steganographic text with controllable semantics is a challenge that needs to be solved.

Automaticlly generating specific semantic text has been a very important research topic in the field of natural language processing for a long time. It is directly related to many valuable research topics, such as automatic image captioning \cite{yang2017image}, automatic translation \cite{Bahdanau2014Neural}, automatic dialogue generation \cite{li2016deep}, etc. Most of these models follow the unified technical framework, which is called Encoder-Decoder framework. They use a specific encoder to encode the semantic information that needs to be expressed (like an image in the image captioning task) into a semantic vector, and then send it to the decoder, and the decoder then generates natural text containing this specific semantic. However, in order to ensure the universality of steganography algorithm, we usually assume that the embedded secret information is arbitrary (without specific semantics). Under this premise, if we further require the generated steganographic sentence to contain specific semantics, it is thus extremely challenging.

The existing text generation steganography model framework mainly implements the secret information hiding through coding the conditional probability of each word during the generation process. This technical framework can hardly control the semantic of the generated steganographic sentence. Therefore, in this paper, we consider information hiding on the encoder side without altering the decoder side, so as to control the semantic expression of the generated steganographic sentences to a certain extent.

\section{The Proposed Method}

\subsection{Notations and Problem Statement}

In this paper, we adopt the following notation conventions. Random variables will be denoted by captial letters (e.g., $X$), and their individual values will be denoted by the respective lower case letters (e.g., $x$). The domains over which random variables are denoted by script letters (e.g., $\mathcal{X}$) and the number of all possible values in $\mathcal{X}$ will be denoted as $|\mathcal{X}|$. A graph contains vertices which represent concepts, and edges which represent the relations between vertices they connnect. We define a graph with $N$ vertices as follows:

\begin{equation}
\left\{\begin{array}{l}
G = \{V \in \mathcal{V},E \in \mathcal{E} \subseteq (\mathcal{V} \times \mathcal{V})\},\\
\mathcal{V} = \cup^N_{i=1}\{v_i\}, \quad \mathcal{E} = \cup\{e_{i,j}\}|_{i,j \in [1,N]}, \quad e_{i,j}: v_i \to v_j.
\end{array} 
\right.
\end{equation}

\noindent where $\mathcal{V}$ and $\mathcal{E}$ denotes the vertices set and edges set, respectively. We define the set of all the edges starting with $v_i$ as $\mathcal{E}^i_{out}$, and the set of all the edges ending with $v_j$ as $\mathcal{E}^j_{in}$, that is:

\begin{equation}
\left\{\begin{array}{l}
\forall i,j\in [1,N], \quad e_{i,j}: v_i \to v_j,\\
\mathcal{E}^i_{out} = \cup\{e_{i,j}\}|_{j \in [1,N]}, \quad \mathcal{E}^j_{in} = \cup\{e_{i,j}\}|_{i \in [1,N]}.\\
\end{array} 
\right.
\end{equation}

\noindent We use $|\mathcal{E}^i_{out}|$ and $|\mathcal{E}^i_{in}|$ to represent the number of edges with node $v_i$ as the starting point and the ending point, respectively. We express the set of paths from node $v_i$ to node $v_j$ as $\mathcal{L}_{v_i:v_j}$, and the number of possible paths as $|\mathcal{L}_{v_i:v_j}|$.

In this paper, the proposed Graph-Stega model also uses Encoder-Decoder framework. We try to introduce a knowledge graph at the Encoder side, and then reasonably encode the paths between the nodes in the graph. Then according to the secret information that needs to be embedded, we extract its corresponding path and construct a subgraph, then use a graph embedding network to extract its semantics, and finally send it to the Decoder to generate steganographic text. The overall framework is shown in Figure 1. 

Suppose, there has a secret message set $\mathcal{M}$, a secret key set $\mathcal{K}$ and a graph space $\mathcal{G}$. Therefore, in fact, we want to complete the mapping process from the secret information space $\mathcal{M}$ to the graph space $\mathcal{G}$ and then to the text space $\mathcal{S}$, that is:

\begin{equation}
\left\{\begin{array}{l}
 Emb: \mathcal{M} \times \mathcal{K} \times \mathcal{G} \to \mathcal{S}, f(M,K,G) = S,\\
 Ext: \mathcal{S} \times \mathcal{K} \to \mathcal{M}, f(K,S) = M.\\
             \end{array}  
        \right.
\end{equation}

\begin{figure}[ht]
  \centering
  \includegraphics[width=\linewidth]{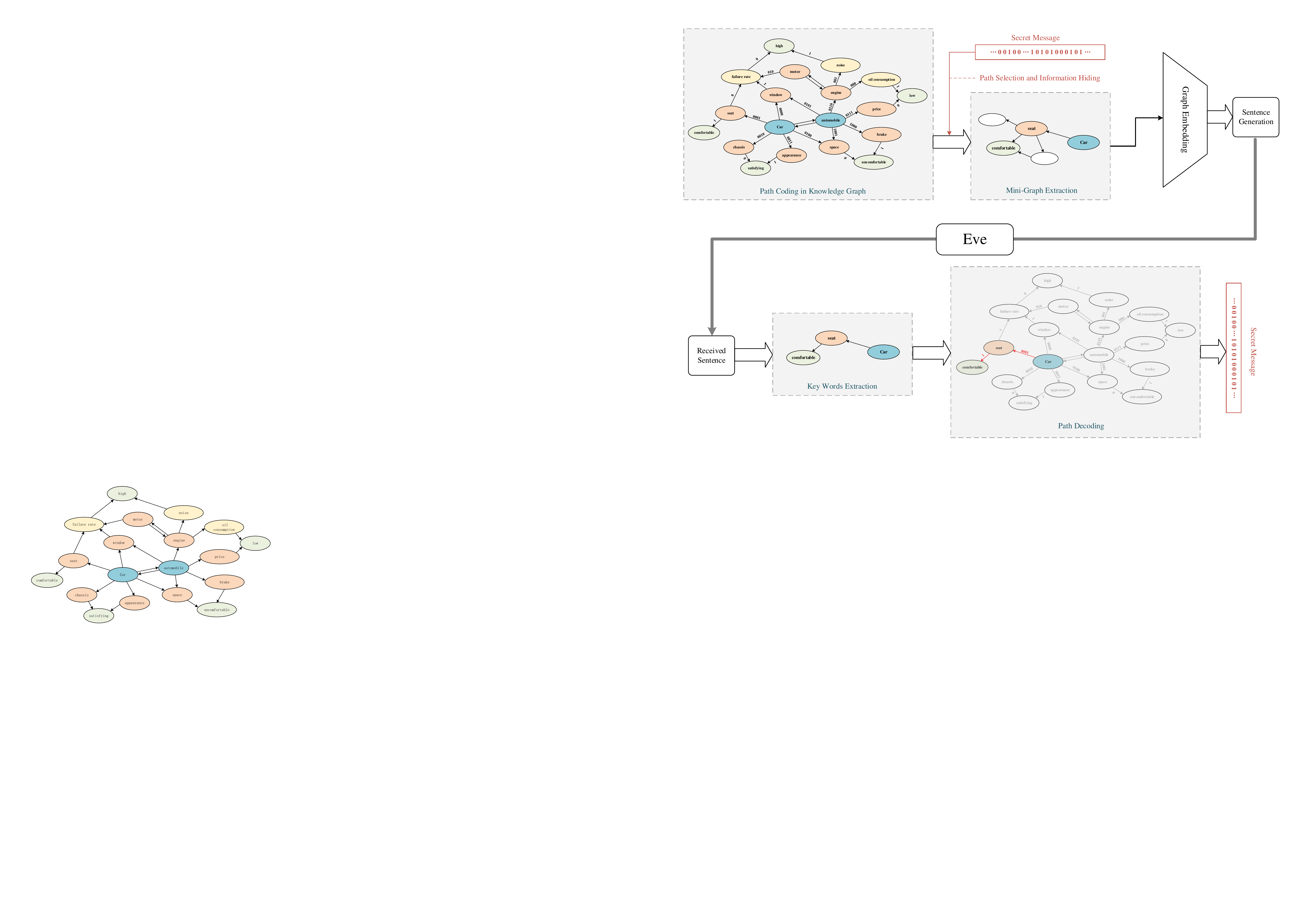}
  \caption{The overall framework of the proposed Graph-Stega model.}
  \label{fig:1}
\end{figure}

\subsection{Path Coding in Knowledge Graph}

The core idea of the proposed method is that, for a given knowledge graph, there are multiple connected paths between any two nodes (e.g., $\mathcal{L}_{v_i:v_j}$ for $v_i$ and $v_j$), each path constitutes a subgraph. Therefore, we can first code these subgraphs reasonably so that different subgraphs (paths) represent different secret information. According to the secret information that needs to be embedded, we then extract the corresponding subgraph from the graph space. Next, we use a specific algorithm to extract the semantic information contained in the subgraph and embedded into a semantic vector $\bm{z}$. Finelly, we use a decoder to generate corresponding steganographic text based on this inputted semantic vector $\bm{z}$. 

For any node, such as $v_i$, there are $|\mathcal{E}^i_{out}|$ edges starting from it, and each edge represents a possible semantic trend. We can convert the encoding of the subgraph to the encoding of the edge set from each node. In this paper, we use Huffman tree to encode the edge set from each node, and the weight is the frequency of each edge appearing in the whole corresponding data set. This enables us to code the semantic trend of each node. The next question is, how do we choose the starting and ending nodes of the path? Obviously this cannot be chosen randomly, otherwise it will increase the difficulty of the decoder to generate natural text.

Here, we consider the fact that there always be a hierarchical structure in a knowledge graph. Some nodes in the knowledge graph represent an objective entity, such as a car, while other nodes may represent the first-level attributes corresponding to a specific entity, such as an engine, and some nodes represent their corresponding second-level attributes, such as fuel consumption. Here, we divide the nodes in the knowledge graph into different sets according to the semantic level. Assuming that the entire knowledge graph contains $K$ semantic levels, we use $\mathcal{V}^k$ to represent the set of nodes in $k$-th semantic level, and the number of nodes in $k$-th semantic level is denoted as $|\mathcal{V}^k|$. Then, from the nodes in $\mathcal{V}^1$ to the nodes in $\mathcal{V}^K$ constitutes a semantic Markov chain, that is:

\begin{equation}
\mathcal{V}^1 \to ... \to \mathcal{V}^k \to ... \to \mathcal{V}^K, k\in [1,K].
\end{equation}

\noindent Therefore, we can specify that during the path encoding process, for each edge, the semantic level of the ending node will not be higher than or equal to the semantic level of the starting node. Under this restriction, each path will correspond to a relatively clear and complete semantic information. For example, for a path like ``automobile $\to$ engine $\to$ fuel consumption", the semantic scope of this path is basically clear, but at the same time, the corresponding natural text still has a certain degree of freedom. For example, the corresponding text can be ``fuel consumption is a normal phenol with internal combination engines in cars", or just ``I don't like this car very much because the fuel consumes of the engine is too much". 

For Alice, if she wants to control the semantics of the generated steganographic sentence, she can only code other nodes and paths by fixing specific words in the graph. For example, if she wants to generate steganographic sentence which describes the engine, she can fix the ``engine" node. Or if Alice wants to generate sentence with positive sentiment, she can fix the ``good" node. In this way, Alice can control the semantic of generated steganographic sentences to a certain extent by sacrificing a little embedding rate (reducing the freedom of some nodes).

\subsection{Subgraph Embedding}

After extracting subgraphs from the knowledge graph based on the secret information, we need to further extract their semantics and convert the subgraphs into corresponding semantic vectors, and then send them into the decoder to generate natural texts. In this paper, we mainly refer to the model proposed in \cite{song2018graph} for semantic extraction of subgraphs.

Firstly, we can express each edge, like $e_{i,j}$, in the graph as a triple $< v_i, v_j, l_{v_i:v_j} >$, where $v_i$ represents the starting node, $v_j$ represents the ending node, and $l_{v_i:v_j}$ represents the connection relationship of them. Secondly, we can map the words represented by each node and each edge into word vector form , such as using $h_i$ represent $v_i$, and $x_{i,j}$ represent $l_{v_i:v_j}$. Then, we use a recurrent neural network with GRUs to update the vectorized semantic representation of nodes and edges according to the information flow in the subgraph. The specific update strategy for $h_i$ and $x_{i,j}$ is as follows:

\begin{flalign}
& \left\{\begin{array}{l}
 I^t_i = \sigma(W_i \cdot [h^{t-1}_i,x^{t-1}_{i,j}] + b_i), \\
 F^t_i = \sigma(W_f \cdot [h^{t-1}_i,x^{t-1}_{i,j}] + b_f),\\
 C^t_i = F^t_i \cdot C^{t-1}_i + I^t_i \cdot \tanh(W_c \cdot [h^{t-1}_i,x^{t-1}_{i,j}] + b_c),\\
 O^t_i = \sigma(W_o \cdot [h^{t-1}_i,x^{t-1}_{i,j}] + b_o),\\
 h^t_i = O^t_i \cdot \tanh(C^t_i),\\
 x^t_{i,j} = W_l \cdot [h^t_i,x^{t-1}_{i,j}] + b_l.\\
             \end{array}  
        \right.&
\end{flalign}

\noindent Where $I^t_i$, $F^t_i$, $O^t_i$ indicate the input gate, the forget gate and the output gate at $t$-th step, respectively. $W_.$ are the weights in them and $b_.$ are the bias.  

Considering the contribution of each node and edge to the semantic representation of the whole graph may be different. Therefore, in order to form a complete semantic representation of the input subgraph, we use attention mechanism \cite{Bahdanau2014Neural} to fuse the weighted information of the last iteration node vectors and semantic vectors, that is:

\begin{equation}
\begin{aligned}
&\alpha_{i} = \frac{exp(\phi(x^t_{i,j},h^t_i))}{\sum_{(i,j,t)}exp(\phi(x^t_{i,j},h^t_j))},\\ 
&\bm{z} = \sum_{(i,j,t)} \alpha_{t} \cdot [x^t_{i,j};h^t_j]
\end{aligned}
\end{equation}

\noindent Here, $\phi()$ represents a fully connected neural network layer. Finally, the semantic information of the whole graph is contained in the semantic vector $\bm{z}$.

\subsection{Sentence Generation}

For a text set with a dictionary $D$, each sentence $S$ with length $n$ is sampled out from the space $\mathcal{C} = \{w_i|w_i \in D\}^n$. However, most combinations do not contain complete semantic information. In order to obtain a semantically complete word sequence, the most common approach is based on statistical language model \cite{Bengio2003A}. Statistical Language Model (LM) first learns conditional distribution probability of each word in normal sentences by training on a large normal sentences set, that is:

\begin{equation}
\begin{aligned}
p(S) & = p(w_1,w_2,w_3,...,w_n)\\
& = p(w_1)p(w_2\mid w_1)...p(w_n\mid w_1,w_2,...,w_{n-1}),
\end{aligned}
\end{equation}

\noindent where $S$ denotes the whole sentence and $w_i$ denotes the $i$-th word in it. The task of decoder is to find a suitable word sequence with complete semantics and correct syntax among $N^n$ possible combinations according to the semantic vector $\bm{z}$. In this work, we use recurrent neural network with LSTM units \cite{Hochreiter1997Long} as the decoder. Its mathematical description is much similar to formula (5), and for simplicity, we denote the transfer function of LSTM units by $f_{LSTM}(\ast)$. RNN can learn the statistical language model from a large number of normal texts, and then calculate the conditional probability distribution of the next word according to the previous generated words, and finally it can generate sentences that conform to such statistical language model. 

For example, supporse currently we have alread generate $i-1$ words and given semantic vector $z$, then the model will calculate the probability distribution of the $i$-th word:

\begin{equation}
p(w_i) = f_{LSTM}(w_1,w_2,...,w_{i-1}|\bm{z})
\end{equation}

The previous steganographic text generative model mainly encodes this conditional probability distribution of each word and to embed the secret information \cite{yang2018rnn,yang2018rits,ziegler2019neural,dai2019towards}. But as we mentioned before, their common problem is that, with the increase of embedding rate, the model will gradually select words with lower conditional probability, thus reducing the quality of generated text. The proposed model mainly conduct steganography in the knowledge graph at the encoder side, so we do not need to modify the conditional probability of each word in the decoder, and we can choose the word with the highest probability as the current output every time, so as to ensure the quality of the generated text.

To ensure that Bob can decode successfully, we need to ensure that the generated text contains the nodes in the subgraph. To solve this challenge, we refer to \cite{gulcehre2016pointing,gu2016incorporating} and introduce the copy mechanism into the text generation process. This mechanism calculates the final vocabulary distribution from two parts:

\begin{equation}
p(w_i) = \theta_ip(w_i\in D)+(1-\theta_i)p(w_i\in G).
\end{equation}

\noindent Where $\theta_i$ is a switch for controlling generating a word from the vocabulary $D$ or directly copying it from the input graph. By introducing this mechanism, we find that most of the generated text can contain the words of the nodes in the input subgraph. In fact, in our task of covert communication, even if the generated text does not contain specific words, Alice can regenerate it again until the sentence contains the required words (the probability of this situation is less than 5\% in our experiments).

When Bob receives the sentences transmitted from Alice, the corresponding path of each sentence in the knowledge map is unique, so it can ensure that Bob can accurately extract the hidden information.

\section{Model Analysis and Experiments}

\subsection{Dataset and Model Training}

Before using the proposed method for covert communication, Alice and Bob are required to select a publicly available knowledge map in advance. In this paper, we use the automobile review dataset and corresponding knowledge graph constructed in \cite{chen2019knowledge} to verify and test the proposed steganography algorithm. This knowledge graph is stored in the form of triples, that is, there are three semantic levels of nodes: entity $\to$ attribute $\to$ sentiment. The whole knowledge graph contains 36,373 triples, corresponding to more than 100,000 natural sentences.

We train the proposed model using maximum likelihood with a regularization term on the attention weights by minimizing a loss function over training set. The loss function is a negative log probability of the ground truth words, that is:

\begin{equation}
Loss = -\sum\log(p(w_i))+\lambda\sum(1-\sum\alpha_{i})^2.\\
\end{equation}

\noindent Where $w_i$ is the ground truth word and $\lambda>0$ is a balancing factor between the cross entropy loss and a penalty on the attention weights. We use stochastic gradient descent with momentum 0.9 to train the parameters of our netwok.

\subsection{Hidden Capicity and Semantic Correlation}

The existing steganography methods based on conditional probability coding mainly control the embedding rate by adjusting the size of candidate pool of each word. The larger the candidate pool is, the higher the embedding rate is, but at the same time, the more likely to select words with lower conditional probability,. Therefore, this kind of method will form a mutual restriction relationship between the information embedding rate and the quality of the generated text. In the proposed model, we encode the path and embed the secret information in the knowledge graph, and transform the constraint into the constraint between the information embedding rate and the semantic controllability, so that the embedding rate will not affect the text quality.

The calculation method of embedding rate is to divide the actual number of embedded bits by the number of bits occupied by the entire generated text in the computer. For the proposed method, the information embedding rate can be expressed as follows:

\begin{equation}
\begin{aligned}
ER = \frac{1}{N}\sum^N_{i=1}\frac{\sum^{K_i}_{k=1}|\mathcal{E}^k_{out}|}{8 \times \sum^{L_j}_{j=1}m_{i,j}} = \frac{\sum^{K_i}_{k=1}|\mathcal{E}^k_{out}|}{8 \times \overline{L} \times \overline{m}},
\end{aligned}
\end{equation}

\noindent where $N$ is the number of generated sentences and $L_i$ is the length of $i$-th sentence. The denominator indicates the number of bits occupied by the $i$-th sentence in the computer. Since each English letter actually occupies one byte in the computer, i.e. 8 bits, the number of bits occupied by each English sentence is $B(s_i) = 8 \times \sum^{L_i}_{j=1}m_{i,j}$, where $m_{i,j}$ represents the number of letters contained in the $j$-th word of the $i$-th sentence. $\overline{L}$ and $\overline{m}$ represent the average length of each sentence in the generated text and the average number of letters contained in each word. $T_i$ represents the length of the semantic chain (formula (4)) corresponding to the $i$-th generated text, and $|\mathcal{E}^k_{out}|$ represents the number of edges starting from the $i$-th node. Alice can select some nodes in the fixed semantic chain to control the semantics of the generated hidden text, but it will lose some of the final information embedding rate.

During the experiment, we adjusted the embedding rate by fixing one node, two nodes, and non-fixed nodes in the triplet to generate steganographic sentences with different embedding rates. Then, we further tested the semantic association between the generated steganographic sentences and the input subgraph. For each input subgraph, we derive its corresponding text from the dataset as our standard reference. Then by comparing the sentence we generated, we calculated several standard metrics used in automatic translation tasks: BLEU \cite{papineni2002bleu}, METEOR \cite{denkowski2014meteor}, CIDEr \cite{vedantam2015cider} and ROUGH-L \cite{lin2006information} (the higher the better). Under different embedding rates, the test results of semantic relevance between the generated steganography and the input subgraph are shown in Table 1.

\begin{table}[h]
\renewcommand\arraystretch{1.2}
\centering
\caption{\label{tab:1}The semantic relevance between the generated steganography and the input subgraph.}
\begin{tabular}{l|c|c|c|c|c}
\toprule[1.5pt]
bpw &BLUE-1 &BLUE-2 &METEOR &CIDEr &ROUGE-L\\
\hline
1.23 &0.453 &0.146 &0.231 &1.184 &0.446\\
\hline
2.01 &0.389 &0.120 &0.229 &1.065 &0.452\\
\hline
2.49 &0.365 &0.109 &0.224 &0.945 &0.460\\
\bottomrule[1.5pt] 
\end{tabular}
\end{table}

According to the test results, we found that the generated steganography text can obey the input semantic information to a certain extent. And under different embedding rates, its semantic relevance will not change much.

\subsection{Quality Evaluation}

Furthermore, we want to know whether the quality of the steganographic sentences generated by the proposed method are reliable, and also, whether it will decline significantly with the increase of embedding rate. In the field of Natural Language Processing, $perplexity$ is a standard metric for sentence quality testing \cite{fang2017generating,yang2018rnn}. It is defined as the average per-word log-probability on the test texts:

\begin{equation}
perplexity = 2^{-\frac{1}{n}\sum^n_{j=1}log{p(w_i\mid w_1,w_2,...,w_{j-1})}},
\end{equation}

\noindent where $s = \{w_1,w_2,w_3,...,w_n\}$ is the generated sentence, $n$ is the number of words in the generated sentence $s$. Usually, the smaller the value of $perplexity$, the better the language model of the sentences, which indicates better quality the generated sentences. We tested the mean $perplexity$ values of the steganographic sentences generated under different embedding rates and the training sentences in the data set. The results are shown in Table 2.

\begin{table}[h]
\renewcommand\arraystretch{1.2}
\centering
\caption{\label{tab:1}The mean $perplexity$ values of the steganographic sentences generated under different embedding rates and the training sentences.}
\begin{tabular}{l|c|c|c|c}
\toprule[1.5pt]
bpw &1.23 &2.01 &2.49 &Normal Sentences\\
\hline
ppl &180.38 &172.96 &186.05 &164.53\\
\bottomrule[1.5pt] 
\end{tabular}
\end{table}

From the results in Table 2, we can see that the $perplexity$ of generated steganographic sentences is close to that of training sentences, and will not decline significantly with the increase of embedding rate. This further proves that our steganography method has more advantages than the current text generative steganographic methods.

In addition, we also tested the anti detection ability of our generated steganographic sentences. We tried to use the text steganalysis algorithm proposed in \cite{yang2019fast} to detect the generated steganography from normal sentences. When the $bpw = 2.49$, the detection results have been shown in Table 3. From the test results, we found that our model has a certain anti detection ability.

\begin{table}[h]
\renewcommand\arraystretch{1.2}
\centering
\caption{\label{tab:1}Steganalysis results of the generated steganographic sentences.}
\begin{tabular}{l|c|c|c}
\toprule[1.5pt]
bpw &Accuracy &Precision &F1-score\\
\hline
score &0.720 &0.647 &0.775\\
\bottomrule[1.5pt] 
\end{tabular}
\end{table}

\section{Conclusion}

In this paper, we proposed a new text generation based steganographic method. The proposed method abandon the current text steganography framework of ``language model + conditional probability coding", and try to use Knowledge Graph (KG) to guide the generation of steganographic sentences. The experimental results show that the proposed model is effective and has many outstanding characteristics that the current text generation steganography algorithm does not have. We hope that this paper will serve as a reference guide for the researchers to facilitate the design and implementation of better text steganography.

\ifCLASSOPTIONcaptionsoff
  \newpage
\fi



\bibliographystyle{IEEEtran}
%
\bibliography{sample}

\begin{thebibliography}{10}
\providecommand{\url}[1]{#1}
\csname url@samestyle\endcsname
\providecommand{\newblock}{\relax}
\providecommand{\bibinfo}[2]{#2}
\providecommand{\BIBentrySTDinterwordspacing}{\spaceskip=0pt\relax}
\providecommand{\BIBentryALTinterwordstretchfactor}{4}
\providecommand{\BIBentryALTinterwordspacing}{\spaceskip=\fontdimen2\font plus
\BIBentryALTinterwordstretchfactor\fontdimen3\font minus
  \fontdimen4\font\relax}
\providecommand{\BIBforeignlanguage}[2]{{%
\expandafter\ifx\csname l@#1\endcsname\relax
\typeout{** WARNING: IEEEtran.bst: No hyphenation pattern has been}%
\typeout{** loaded for the language `#1'. Using the pattern for}%
\typeout{** the default language instead.}%
\else
\language=\csname l@#1\endcsname
\fi
#2}}
\providecommand{\BIBdecl}{\relax}
\BIBdecl

\bibitem{das2019enabling}
D.~Das, S.~Maity, B.~Chatterjee, and S.~Sen, ``Enabling covert body area
  network using electro-quasistatic human body communication,''
  \emph{Scientific reports}, vol.~9, no.~1, p. 4160, 2019.

\bibitem{Seife917}
\BIBentryALTinterwordspacing
C.~Seife, ``Digital music safeguard may need retuning,'' \emph{Science}, vol.
  290, no. 5493, pp. 917--919, 2000. [Online]. Available:
  \url{https://science.sciencemag.org/content/290/5493/917.2}
\BIBentrySTDinterwordspacing

\bibitem{Voss826}
\BIBentryALTinterwordspacing
D.~Voss, ``Music industry strikes sour note for academics,'' \emph{Science},
  vol. 292, no. 5518, pp. 826--827, 2001. [Online]. Available:
  \url{https://science.sciencemag.org/content/292/5518/826}
\BIBentrySTDinterwordspacing

\bibitem{zhao1996watermarking}
J.~Zhao, ``Watermarking by numbers,'' \emph{Nature}, vol. 384, no. 6609, p.
  514, 1996.

\bibitem{Simmons1984The}
G.~J. Simmons, ``The prisoners’ problem and the subliminal channel,''
  \emph{Advances in Cryptology Proc Crypto}, pp. 51--67, 1984.

\bibitem{moulin2003information}
P.~Moulin and J.~A. O'Sullivan, ``Information-theoretic analysis of information
  hiding,'' \emph{IEEE Transactions on information theory}, vol.~49, no.~3, pp.
  563--593, 2003.

\bibitem{fridrich2009steganography}
J.~Fridrich, \emph{Steganography in digital media: principles, algorithms, and
  applications}.\hskip 1em plus 0.5em minus 0.4em\relax Cambridge University
  Press, 2009.

\bibitem{huang2011steganography}
Y.~F. Huang, S.~Tang, and J.~Yuan, ``Steganography in inactive frames of voip
  streams encoded by source codec,'' \emph{IEEE Transactions on information
  forensics and security}, vol.~6, no.~2, pp. 296--306, 2011.

\bibitem{yang2018rnn}
Z.~Yang, X.~Guo, Z.~Chen, Y.~Huang, and Y.-J. Zhang, ``Rnn-stega: Linguistic
  steganography based on recurrent neural networks,'' \emph{IEEE Transactions
  on Information Forensics and Security}, 2018.

\bibitem{shanableh2012data}
T.~Shanableh, ``Data hiding in mpeg video files using multivariate regression
  and flexible macroblock ordering,'' \emph{IEEE transactions on information
  forensics and security}, vol.~7, no.~2, pp. 455--464, 2012.

\bibitem{michel2011quantitative}
J.-B. Michel, Y.~K. Shen, A.~P. Aiden, A.~Veres, M.~K. Gray, J.~P. Pickett,
  D.~Hoiberg, D.~Clancy, P.~Norvig, J.~Orwant \emph{et~al.}, ``Quantitative
  analysis of culture using millions of digitized books,'' \emph{science}, vol.
  331, no. 6014, pp. 176--182, 2011.

\bibitem{Wayner1992Mimic}
P.~Wayner, ``Mimic functions,'' \emph{Cryptologia}, vol.~16, no.~3, pp.
  193--214, 1992.

\bibitem{chapman1997hiding}
M.~Chapman and G.~Davida, ``Hiding the hidden: A software system for concealing
  ciphertext as innocuous text,'' in \emph{International Conference on
  Information and Communications Security}.\hskip 1em plus 0.5em minus
  0.4em\relax Springer, 1997, pp. 335--345.

\bibitem{moraldo2014approach}
H.~H. Moraldo, ``An approach for text steganography based on markov chains,''
  \emph{arXiv preprint arXiv:1409.0915}, 2014.

\bibitem{dai2010text}
W.~Dai, Y.~Yu, Y.~Dai, and B.~Deng, ``Text steganography system using markov
  chain source model and des algorithm.'' \emph{JSW}, vol.~5, no.~7, pp.
  785--792, 2010.

\bibitem{Luo2016Text}
Y.~Luo, Y.~Huang, F.~Li, and C.~Chang, ``Text steganography based on ci-poetry
  generation using markov chain model,'' \emph{Ksii Transactions on Internet \&
  Information Systems}, vol.~10, no.~9, pp. 4568--4584, 2016.

\bibitem{yang2018rits}
Z.~Yang, P.~Zhang, M.~Jiang, Y.~Huang, and Y.-J. Zhang, ``Rits: Real-time
  interactive text steganography based on automatic dialogue model,'' in
  \emph{International Conference on Cloud Computing and Security}.\hskip 1em
  plus 0.5em minus 0.4em\relax Springer, 2018, pp. 253--264.

\bibitem{fang2017generating}
T.~Fang, M.~Jaggi, and K.~Argyraki, ``Generating steganographic text with
  lstms,'' \emph{arXiv preprint arXiv:1705.10742}, 2017.

\bibitem{dai2019towards}
F.~Z. Dai and Z.~Cai, ``Towards near-imperceptible steganographic text,''
  \emph{arXiv preprint arXiv:1907.06679}, 2019.

\bibitem{ziegler2019neural}
Z.~M. Ziegler, Y.~Deng, and A.~M. Rush, ``Neural linguistic steganography,''
  \emph{arXiv preprint arXiv:1909.01496}, 2019.

\bibitem{Chotikakamthorn1998Electronic}
N.~Chotikakamthorn, ``Electronic document data hiding technique using
  inter-character space,'' in \emph{Circuits and Systems, 1998. IEEE APCCAS
  1998. The 1998 IEEE Asia-Pacific Conference on}, 1998, pp. 419--422.

\bibitem{Xiang2014Linguistic}
L.~Xiang, X.~Sun, G.~Luo, and B.~Xia, ``Linguistic steganalysis using the
  features derived from synonym frequency,'' \emph{Multimedia Tools and
  Applications}, vol.~71, no.~3, pp. 1893--1911, 2014.

\bibitem{shniperov2016text}
A.~Shniperov and K.~Nikitina, ``A text steganography method based on markov
  chains,'' \emph{Automatic Control and Computer Sciences}, vol.~50, no.~8, pp.
  802--808, 2016.

\bibitem{yang2018automatically}
Z.~Yang, S.~Jin, Y.~Huang, Y.~Zhang, and H.~Li, ``Automatically generate
  steganographic text based on markov model and huffman coding,'' \emph{arXiv
  preprint arXiv:1811.04720}, 2018.

\bibitem{yang2019behavioral}
Z.~Yang, Y.~Hu, Y.~Huang, and Y.~Zhang, ``Behavioral security in covert
  communication systems,'' \emph{arXiv preprint arXiv:1910.09759}, 2019.

\bibitem{yang2017image}
Z.~Yang, Y.-J. Zhang, S.~ur~Rehman, and Y.~Huang, ``Image captioning with
  object detection and localization,'' in \emph{International Conference on
  Image and Graphics}.\hskip 1em plus 0.5em minus 0.4em\relax Springer, 2017,
  pp. 109--118.

\bibitem{Bahdanau2014Neural}
D.~Bahdanau, K.~Cho, and Y.~Bengio, ``Neural machine translation by jointly
  learning to align and translate,'' \emph{Computer Science}, 2014.

\bibitem{li2016deep}
J.~Li, W.~S. Monroe, A.~Ritter, D.~Jurafsky, M.~Galley, and J.~Gao, ``Deep
  reinforcement learning for dialogue generation,'' pp. 1192--1202, 2016.

\bibitem{song2018graph}
L.~Song, Y.~Zhang, Z.~Wang, and D.~Gildea, ``A graph-to-sequence model for
  amr-to-text generation,'' \emph{arXiv preprint arXiv:1805.02473}, 2018.

\bibitem{Bengio2003A}
Y.~Bengio, P.~Vincent, and C.~Janvin, ``A neural probabilistic language
  model,'' \emph{Journal of Machine Learning Research}, vol.~3, no.~6, pp.
  1137--1155, 2003.

\bibitem{Hochreiter1997Long}
S.~Hochreiter and J.~Schmidhuber, ``Long short-term memory,'' \emph{Neural
  Computation}, vol.~9, no.~8, pp. 1735--1780, 1997.

\bibitem{gulcehre2016pointing}
C.~Gulcehre, S.~Ahn, R.~Nallapati, B.~Zhou, and Y.~Bengio, ``Pointing the
  unknown words,'' \emph{arXiv preprint arXiv:1603.08148}, 2016.

\bibitem{gu2016incorporating}
J.~Gu, Z.~Lu, H.~Li, and V.~O. Li, ``Incorporating copying mechanism in
  sequence-to-sequence learning,'' \emph{arXiv preprint arXiv:1603.06393},
  2016.

\bibitem{chen2019knowledge}
F.~Chen and Y.~Huang, ``Knowledge-enhanced neural networks for sentiment
  analysis of chinese reviews,'' \emph{Neurocomputing}, vol. 368, pp. 51--58,
  2019.

\bibitem{papineni2002bleu}
K.~Papineni, S.~Roukos, T.~Ward, and W.-J. Zhu, ``Bleu: a method for automatic
  evaluation of machine translation,'' in \emph{Proceedings of the 40th annual
  meeting on association for computational linguistics}.\hskip 1em plus 0.5em
  minus 0.4em\relax Association for Computational Linguistics, 2002, pp.
  311--318.

\bibitem{denkowski2014meteor}
M.~Denkowski and A.~Lavie, ``Meteor universal: Language specific translation
  evaluation for any target language,'' in \emph{Proceedings of the ninth
  workshop on statistical machine translation}, 2014, pp. 376--380.

\bibitem{vedantam2015cider}
R.~Vedantam, C.~Lawrence~Zitnick, and D.~Parikh, ``Cider: Consensus-based image
  description evaluation,'' in \emph{Proceedings of the IEEE conference on
  computer vision and pattern recognition}, 2015, pp. 4566--4575.

\bibitem{lin2006information}
C.-Y. Lin, G.~Cao, J.~Gao, and J.-Y. Nie, ``An information-theoretic approach
  to automatic evaluation of summaries,'' in \emph{Proceedings of the main
  conference on Human Language Technology Conference of the North American
  Chapter of the Association of Computational Linguistics}.\hskip 1em plus
  0.5em minus 0.4em\relax Association for Computational Linguistics, 2006, pp.
  463--470.

\bibitem{yang2019fast}
Z.~Yang, Y.~Huang, and Y.-J. Zhang, ``A fast and efficient text steganalysis
  method,'' \emph{IEEE Signal Processing Letters}, vol.~26, no.~4, pp.
  627--631, 2019.

\end{thebibliography}

\end{document}